%% file: main.tex
\documentclass[10pt,twocolumn,letterpaper]{article}

\usepackage{cvpr}


\definecolor{cvprblue}{rgb}{0.21,0.49,0.74}
\usepackage[pagebackref,breaklinks,colorlinks,allcolors=cvprblue]{hyperref}

\def\paperID{*****}
\def\confName{CVPR}
\def\confYear{2026}

\usepackage{booktabs}
\usepackage{amsmath}
\usepackage{amssymb}
\usepackage{microtype}

\title{SMART: SMPLest-X Mesh Adaptation and RAFT Tracking\\
for Soccer Pose Estimation}

\author{
  Parthsarthi Rawat \\
  GameChanger by Dick's Sporting Goods \\
  {\tt\small sarthi.rawat@gc.com}
}

\begin{document}
\maketitle

\begin{abstract}
We present our approach to the FIFA Skeletal Tracking Challenge 2026, which
requires estimating 3D world-space poses of soccer players from broadcast
video.
Our method fine-tunes \textbf{SMPLest-X} (ViT-H, 687\,M parameters) via
a stratified clip split, multi-task depth supervision, and broadcast
augmentation, paired with a \textbf{RAFT} dense optical flow camera
tracker, foot-plane anchoring, and two-pass temporal smoothing.
Against the FIFA baseline score of 1.053 on the validation set, SMART
achieves \textbf{0.647}, a \textbf{38.6\%} improvement; on the
held-out test set, SMART scores \textbf{0.593}
(Global MPJPE: 0.324\,m, Local MPJPE: 0.054\,m).
\end{abstract}

\section{Task Definition}

Given a broadcast soccer video clip with $T$ frames and $N$ tracked
players, the goal is to predict per-frame, per-player 3D joint positions
in a shared world coordinate system.
Each player is represented by a \textbf{FIFA-15 skeleton}: 15 joints
comprising Nose, right/left Shoulder, Elbow, Wrist, Hip, Knee, Ankle,
and Big Toe (indices 0--14).
The \textbf{root joint} is the mid-hip:
$\mathbf{r}_t = \tfrac{1}{2}(\mathbf{j}_{t,7}+\mathbf{j}_{t,8})$.

Performance is measured by two metrics (in metres):
\textbf{Global MPJPE} measures mean per-joint error in absolute world
coordinates, and \textbf{Local MPJPE} measures mean per-joint error
after subtracting the root.
The competition score is
$\mathcal{L} = \text{Global MPJPE} + 5\times\text{Local MPJPE}$,
reflecting the 5$\times$ higher leverage of local pose quality.

\section{Data}

\paragraph{Competition data.}
The benchmark provides 20 sequences from FIFA World Cup 2022 broadcast
video at 50\,fps and 1920$\times$1080 resolution (6 validation, 14 test).
Per-frame player bounding boxes, camera intrinsics $K_t$, radial
distortion coefficients $k_t$, and an initial camera pose $(R_0,t_0)$
are all provided by the competition; $K_t$ varies per frame to account
for continuous zoom changes in broadcast production.
The baseline propagates subsequent per-frame poses $(R_t,t_t)$ via
Lucas-Kanade sparse optical flow on 714 pre-annotated 3D pitch landmarks.

\paragraph{WorldPose fine-tuning data.}
We fine-tune on \textbf{WorldPose}~\cite{jiang2024worldpose}, which pairs
broadcast soccer clips with pseudo-ground-truth 3D body poses.
We partition 89 clips at clip boundaries into a \emph{stratified} split:
70 training and 19 validation clips with zero frame overlap and matched
camera-height and viewing-angle distributions.
No competition frames are used during training.

\section{Methodology}

Our pipeline has four stages:
(1) SMPLest-X recovers each player's camera-space root-relative skeleton
and absolute pelvis depth from their image crop;
(2) RAFT-small propagates per-frame camera rotation $R_t$ and translation
$t_t$ from the provided $(R_0,t_0)$;
(3) the skeleton and pelvis depth are combined with $(R_t,t_t)$ to place
each player in the shared world coordinate system and anchor them to the
pitch plane; and
(4) world trajectories and root-relative joints are independently
smoothed to remove detection jitter.

\subsection{3D Body Mesh Recovery}

We use \textbf{SMPLest-X}~\cite{yin2025smplest}, a regression-based
expressive body mesh recovery model (ViT-H backbone, 687\,M parameters)
built on the \textbf{SMPL-X} body representation --- a parametric model
that encodes human shape and pose as low-dimensional coefficients.
Each player's bounding box is first expanded by $1.2\times$ to include
contextual background, then resized to $512\times384$ pixels via affine
transform.
Given the player crop, SMPLest-X regresses SMPL-X parameters
(shape $\boldsymbol\beta$, pose $\boldsymbol\theta$, global orientation)
and produces a full body mesh ($V{=}10{,}475$ vertices).
FIFA-15 joints are read off this mesh via a fixed per-joint vertex
lookup: each of the 15 competition joints maps to a designated
SMPL-X surface vertex, requiring no additional learned module.

\paragraph{Domain-adaptive fine-tuning.}
Pre-trained SMPLest-X suffers a domain gap on broadcast soccer: small
player size, wide viewing angles, and partial occlusions degrade accuracy.
We address this via a \emph{stratified} clip-boundary split and a
multi-task loss
$\mathcal{L}_\text{train} = \lambda_1\mathcal{L}_\text{3D}
+ \lambda_2\mathcal{L}_\text{2D} + \lambda_3\mathcal{L}_\text{depth}$,
where $\mathcal{L}_\text{3D}$ is weighted MPJPE (extremities at
$3\times$ weight), $\mathcal{L}_\text{2D}$ is L2 on projected keypoints,
and $\mathcal{L}_\text{depth}$ is L1 on camera-space pelvis depth,
the single most important factor for world-space grounding
(see Table~\ref{tab:ablation_ft}).
Broadcast augmentation completes the training recipe: random resized
cropping (scale $\in[0.75,1.0]$, 80\% probability), horizontal flipping
(50\%, with symmetric joint permutation), and colour jitter ($\pm0.4$
brightness/contrast/saturation, $\pm0.1$ hue).

\subsection{Camera Pose Estimation}

With per-frame intrinsics $K_t$ and distortion $k_t$ provided, we need
only propagate camera rotation $R_t\in SO(3)$ and translation
$t_t\in\mathbb{R}^3$ from the initial $(R_0,t_0)$.
We replace the competition baseline's Lucas-Kanade sparse tracker with
\textbf{RAFT-small}~\cite{teed2020raft}, which provides substantially
more reliable correspondences on broadcast grass textures, particularly
during panning and zoom.
Per frame, we compute dense flow from $I_{t-1}$ to $I_t$ and sample it
inside the convex hull of the 714 projected pitch landmarks at stride 4,
yielding ${\sim}50\times$ more correspondences than landmark tracking alone.
Flow vectors contaminated by player motion are removed via
MAD outlier rejection ($3\sigma$ threshold)~\cite{leys2013mad}
before fitting an inter-frame homography via
RANSAC and decomposing to $(R_t,t_t)$; updates with
$|\Delta R|>60^{\circ}$ are rejected to guard against tracking failures.
When field line intersections are detected via Hough transform, an
EPnP (Efficient Perspective-$n$-Point)~\cite{lepetit2009epnp} solve
further corrects absolute camera position on the first frame.

\subsection{World-Space Skeleton Placement}

With per-frame camera poses established, we lift each player's
camera-space skeleton into the shared world coordinate system in
three steps.

\noindent\textbf{(1) Rotation and initial placement.}
The root-relative skeleton is rotated to world frame as
$\hat{\mathbf{S}}_t = R_t\,\mathbf{S}_t^{(\mathrm{cam})}$.
The camera-space pelvis translation $\mathbf{p}_t^{(\mathrm{cam})}$
predicted by SMPLest-X is projected to world space as
$\mathbf{p}_t = R_t\,\mathbf{p}_t^{(\mathrm{cam})} + t_t$,
establishing the player's initial absolute position.

\noindent\textbf{(2) Foot-plane anchoring.}
The lowest foot pixel in the bounding box is back-projected to a
camera ray using $K_t$ and intersected with the known pitch-plane
equation, yielding ground contact point $\mathbf{f}_t$;
all joints are then translated so the lower ankle aligns
with $\mathbf{f}_t$:
\begin{equation}
  \mathbf{J}_{t,j}
    = \hat{\mathbf{S}}_{t,j}
    - \hat{\mathbf{S}}_{t,\mathrm{ankle}}
    + \mathbf{f}_t.
\end{equation}

\noindent\textbf{(3) Reprojection refinement.}
A two-stage L-BFGS optimisation (50 steps translating the skeleton
centroid onto the bounding-box centre, then 50 steps aligning projected
joints with 2D keypoints) refines the absolute placement.

\subsection{Temporal Smoothing}

Raw per-frame predictions exhibit jitter from detection noise and
occasional misassociated crops under occlusion; we apply a two-pass
smoothing strategy to produce clean trajectories.
In the first pass, global root trajectories are cleaned: MAD outlier
detection flags teleporting jumps, which are replaced by linear
interpolation before a Gaussian filter removes residual high-frequency
noise.
In the second pass, root-relative joint offsets are smoothed
independently with a narrower Gaussian to remove single-frame jitter
while preserving rapid limb movements such as kicks; the world centroid
is frozen during this step so the two passes do not interfere.

\section{Implementation Details}
\label{sec:impl}

\paragraph{Fine-tuning.}
Only the SMPLest-X decoder and prediction heads are trainable; the
ViT-H backbone is frozen.
Training: 15 epochs, batch size 16, Adam, LR $2{\times}10^{-6}$,
AMP mixed precision, 1$\times$ 24\,GB A10G GPU.
Loss weights: $\lambda_1{=}1.0$, $\lambda_2{=}0.1$, $\lambda_3{=}0.5$;
per-joint weights $3.0$ for wrists (5,6), ankles (11,12), big toes
(13,14), and $1.0$ otherwise.

\paragraph{Camera tracking.}
RAFT-small (default pretrained weights); stride-4 dense flow inside
the 714-landmark convex hull; MAD $3\sigma$ outlier rejection;
RANSAC homography (2000\,iter, 1\,px threshold);
$|\Delta R|{>}60^{\circ}$ rejection.

\paragraph{Smoothing.}
Global: MAD ($5\times$) + linear interp + Gaussian ($\sigma{=}3$\,fr),
clamp at 0.35\,m/fr.
Local joints: independent Gaussian ($\sigma{=}1.5$\,fr).

\section{Results}

\paragraph{Main results.}
Table~\ref{tab:main} compares our method against the FIFA competition
baseline.
Against the FIFA baseline on the validation set, our method reduces the
competition score by \textbf{38.6\%} (1.053$\to$0.647).
On the held-out test set, SMART achieves \textbf{0.593}, demonstrating
strong generalisation beyond the validation distribution.

\begin{table}[h]
  \centering
  \caption{Results vs.\ the FIFA baseline.}
  \label{tab:main}
  \setlength{\tabcolsep}{5pt}
  \begin{tabular}{lccc}
    \toprule
    Method & Global$\downarrow$ & Local$\downarrow$ & Score$\downarrow$ \\
    \midrule
    FIFA Baseline (val) & 0.602 & 0.090 & 1.053 \\
    \midrule
    Ours (val)  & 0.370 & 0.055 & 0.647 \\
    Ours (test) & \textbf{0.324} & \textbf{0.054} & \textbf{0.593} \\
    \bottomrule
  \end{tabular}
\end{table}

\subsection{Ablation: Fine-tuning Strategy}
\label{sec:abl_ft}

We ablate each fine-tuning design choice incrementally on the validation
split. Results are shown in Table~\ref{tab:ablation_ft}.

\noindent\textbf{FIFA Baseline (SAM-3D).}
The competition baseline uses SAM-3D-Body~\cite{yang2026sam3dbody} as
the 3D lifter, scoring 1.053 on the validation split.

\noindent\textbf{SAM-3D + MLP.}
We augmented the baseline with a learned Placement MLP predicting
absolute hip position from camera geometry and foot-ray features
(the ray cast from the camera through the player's lowest visible foot
pixel to the pitch plane), reducing the score to 0.864 (global:
0.470\,m, local: 0.079\,m).
The MLP generalises poorly across the range of camera heights in the
competition sequences, motivating a switch to SMPLest-X.

\noindent\textbf{SMPLest-X pretrained.}
Off-the-shelf SMPLest-X ViT-H (no domain adaptation) achieves score
0.846 (global: 0.522\,m, local: 0.065\,m).
The domain gap with broadcast soccer video limits out-of-the-box
effectiveness, motivating domain-specific fine-tuning.

\noindent\textbf{+ Na\"{i}ve fine-tuning.}
A chronological split with standard 3D MPJPE loss improves global MPJPE
to 0.425\,m (score: 0.820), but local MPJPE \emph{worsens} from 0.065
to 0.079\,m: adjacent clips share nearly identical camera angles, so
the model overfits training-set viewpoints at the cost of held-out
local accuracy.

\noindent\textbf{+ Foot-plane anchoring.}
Without explicit pitch-plane constraints, skeletons float at arbitrary
height, inflating both global and local error.
Ray-casting the lowest foot pixel to the pitch plane reduces global
MPJPE by 44\,mm and local MPJPE by 12\,mm (score: 0.820$\to$0.714),
confirming that grounding improves both absolute placement and
root-relative scale.

\noindent\textbf{+ Improved training (ours).}
Our full strategy adds (i) stratified clip-boundary split, (ii) depth
supervision $\mathcal{L}_\text{depth}$, and (iii) broadcast augmentation,
yielding local 0.055\,m and global 0.370\,m for a final score of 0.647.
Depth supervision is the dominant factor: without it, local MPJPE
stagnates at 0.067\,m regardless of augmentation.

\begin{table}[t]
  \centering
  \caption{Incremental fine-tuning ablation on the validation split.
           From \emph{SMPLest-X pretrained} onward, rows use the RAFT
           camera tracker and foot-ray placement.}
  \label{tab:ablation_ft}
  \small
  \setlength{\tabcolsep}{3pt}
  \begin{tabular}{lccc}
    \toprule
    Configuration & G (m) & L (m) & Score \\
    \midrule
    FIFA Baseline (SAM-3D)       & 0.602 & 0.090 & 1.053 \\
    SAM-3D + MLP                 & 0.470 & 0.079 & 0.864 \\
    SMPLest-X pretrained         & 0.522 & 0.065 & 0.846 \\
    $+$ na\"{i}ve fine-tune    & 0.425 & 0.079 & 0.820 \\
    $+$ foot-plane anchoring   & 0.381 & 0.067 & 0.714 \\
    $+$ improved training (\textbf{ours})  & \textbf{0.370} & \textbf{0.055} & \textbf{0.647} \\
    \bottomrule
  \end{tabular}
\end{table}

\subsection{Ablation: Camera Tracker}
\label{sec:abl_cam}

We compared optical flow methods for camera pose propagation, measuring
rotation error against ground-truth camera poses on our WorldPose
validation split (19 clips; Table~\ref{tab:ablation_cam}).
ECC homography ($0.107^\circ$) performs worst: soccer frames contain many
moving players that occupy a large fraction of the image, and ECC has
no mechanism to ignore them, so their motion corrupts the background
homography estimate causing systematic drift.
Lucas-Kanade sparse tracking ($0.043^\circ$) handles this via RANSAC
inlier rejection of player-located feature points.
RAFT-small without filtering ties Lucas-Kanade, as dense flow over
player silhouettes introduces noise that cancels the correspondence
benefit; adding MAD outlier filtering removes contaminated pixels and
reduces error to $0.041^\circ$.
RAFT-large provides no further gain at double the compute, so we adopt
RAFT-small with MAD outlier filtering.

\begin{table}[t]
  \centering
  \caption{Camera tracker ablation. Rotation error ($^\circ$) vs.\ GT camera
           poses on our WorldPose validation split (19 clips).}
  \label{tab:ablation_cam}
  \small
  \setlength{\tabcolsep}{3pt}
  \begin{tabular}{lcc}
    \toprule
    Tracker & Rot.\ Err ($^\circ$)$\downarrow$ & ms/frame \\
    \midrule
    ECC homography              & 0.107 & --- \\
    Lucas-Kanade (baseline)     & 0.043 & --- \\
    RAFT-small, no filter       & 0.043 & 55 \\
    \textbf{RAFT-small + MAD (ours)} & \textbf{0.041} & \textbf{55} \\
    RAFT-large + MAD            & 0.041 & 110 \\
    \bottomrule
  \end{tabular}
\end{table}

\subsection{Ablation: Temporal Refiner}
\label{sec:abl_refiner}

We trained a sliding-window temporal refinement network (27-frame
window, hidden dim 128) on WorldPose to suppress per-frame jitter.
While it reduces local MPJPE by 22\% on our WorldPose validation split, competition
performance degrades (score 0.680 vs.\ 0.647) as the refiner
over-smooths fast actions not well-represented in its training
distribution.
We therefore use only Gaussian smoothing in our final submission.

\section{Discussion}

\paragraph{Global vs.\ local error.}
Global MPJPE contributes $\sim$55\% of our total competition score,
making camera tracking the primary bottleneck.
Sequences with aggressive zoom cause apparent pitch-plane deformation
that confounds the homography model, raising global error by up to
$2\times$ compared to stable wide-angle shots.
In contrast, local MPJPE is largely camera-independent once the
fine-tuned lifter is in place.

\paragraph{Limitations.}
(i) Diving poses are underrepresented in WorldPose; SMPLest-X
struggles with extreme configurations, and foot-plane anchoring
breaks when the player is airborne.
(ii) SMPLest-X is per-frame only, making it fragile under occlusion;
a temporally-aware lifter would handle occluded frames more robustly.
(iii) The 687\,M ViT-H backbone dominates inference cost; ViT-S
could yield a $4\times$ speedup with modest accuracy loss.

{
\small
\bibliographystyle{ieee_fullname}
\bibliography{main}
}

\end{document}